\documentclass[letterpaper]{article} 
\usepackage{aaai25}  
\usepackage{times}  
\usepackage{helvet}  
\usepackage{courier}  
\usepackage[hyphens]{url}  
\usepackage{graphicx} 
\usepackage{natbib}  
\usepackage{caption} 
\usepackage{algorithm}
\usepackage{algorithmic}

\usepackage{hhline}
\usepackage{multirow}
\usepackage{dsfont}
\usepackage{caption}
\usepackage{subcaption}
\usepackage{textcomp}
\usepackage{lineno}

\usepackage{amsmath,amssymb,amsfonts}%
\usepackage{amsthm}%
\usepackage{mathrsfs}%

\usepackage{adjustbox}

\usepackage{newfloat}
\usepackage{listings}
\DeclareCaptionStyle{ruled}{labelfont=normalfont,labelsep=colon,strut=off} 
\floatstyle{ruled}
\newfloat{listing}{tb}{lst}{}
\floatname{listing}{Listing}
\title{Hybrid Phenology Modeling for Predicting Temperature Effects on Tree Dormancy}
\author{
    Ron van Bree \textsuperscript{\rm 1}, 
    Diego Marcos \textsuperscript{\rm 2}, 
    Ioannis N. Athanasiadis \textsuperscript{\rm 1}
}
\affiliations{
    \textsuperscript{\rm 1}Artificial Intelligence Group, Wageningen University \& Research,\\
    \textsuperscript{\rm 2}Inria, University of Montpellier\\
    ron.vanbree@wur.nl, diego.marcos@inria.fr, ioannis.athanasiadis@wur.nl
}

\newcommand{\modelname}{Hybrid}
\newcommand{\doy}{DoY}

\newcommand{\argmax}{\mathop{\mathrm{argmax}}}

\begin{document}

\maketitle

%
%
\begin{abstract}

Biophysical models offer valuable insights into climate-phenology relationships in both natural and agricultural settings. However, there are substantial structural discrepancies across models which require site-specific recalibration, often yielding inconsistent predictions under similar climate scenarios. Machine learning methods offer data-driven solutions, but often lack interpretability and alignment with existing knowledge. We present a phenology model describing dormancy in fruit trees, integrating conventional biophysical models with a neural network to address their structural disparities. We evaluate our hybrid model in an extensive case study predicting cherry tree phenology in Japan, South Korea and Switzerland. Our approach consistently outperforms both traditional biophysical and machine learning models in predicting blooming dates across years. Additionally, the neural network's adaptability facilitates parameter learning for specific tree varieties, enabling robust generalization to new sites without site-specific recalibration. This hybrid model leverages both biophysical constraints and data-driven flexibility, offering a promising avenue for accurate and interpretable phenology modeling.
    
\end{abstract}

%
\begin{links}
\footnotesize
    \link{Code}{https://github.com/WUR-AI/HybridML-Phenology}
\end{links}

%
%
%
%
\section{Introduction}\label{section:introduction}

Modeling tree phenology is important to understand the effects of a changing climate on natural and agricultural ecosystems. Temperate trees require a sustained period of cold temperatures to overcome dormancy, protecting them from harsh winter conditions. Temperature changes can delay or accelerate this process, making this phenological process an important component in modeling secondary effects, such as impacts on orchard yields~\citep{baldocchi2008accumulated, luedeling2012climate}, biological carbon sequestration~\citep{cleland2007shifting, richardson2010influence} and the phenological stages of other organisms~\citep{visser2001warmer, kudo2014vulnerability}). Cherry trees are of special interest in climate change research, as their flowering dates are strongly temperature dependent~\citep{samish1954dormancy} and their cultural significance in Japan has allowed the collection of a long record of flowering dates throughout the country (going back $\sim$1200 years in Kyoto~\citep{aono2008phenological}). Trends in cherry tree blooming dates data have been used as an indicator of changes in climate~\citep{aono1993variation, primack2009impact}. 






Biophysical phenology models generally assume the daily accumulation of temperature-dependent units towards some threshold at which a phenological transition occurs. Models typically distinguish between endodormancy, in which a period of cold temperatures needs to be endured, and a subsequent requirement for warmer temperatures during ecodormancy. Commonly used models, however, differ in how they relate daily temperature to their contribution towards endodormancy release and display differences in behaviour under the same climate scenarios~\citep{luedeling2011global, fernandez2020importance}.  The models have typically been proposed as the simplest functions that are able to explain observations from controlled experiments on optimal chilling temperatures and effective temperature ranges, and weight their contribution accordingly. These models work well for predicting blooming dates but do not fully capture the true dynamics that are known to be more complex~\citep{kaufmann2019substitution}. Given the uncertainty in our understanding of these biological processes, choosing a particular model often introduces a strong bias.

The increasing availability of data on climate conditions, along with large datasets on corresponding plant phenology, allows for a more data driven approach to model these dynamics through expressive machine learning methods~\citep{masago2022estimating}. Complex over-parameterized models, due to their tendency to overfit to the training conditions, have difficulty matching simple process-based models based on biological insights when tasked to predict dormancy release based solely on temperature data~\citep{saxena2023multi}. \cite{nagai2020simpler} show self-organizing maps can be used to predict flowering dates, but show higher prediction error than process-based models. Moreover, the inherent interpretability provided by these process-based models is lost~\citep{rudin2019stop}. Introducing the appropriate inductive bias to machine learning models by coupling them with mechanistic models (an approach often referred to as hybrid modelling) is widely recognized as a means to improve their data-efficiency and interpretability by constraining them to solutions that are consistent with domain knowledge and has been successfully applied in a wide variety of domains~\citep{karpatne2017theory,reichstein2019deep, shen2023differentiable}. \\

\begin{figure}[t]
    \centering
    \includegraphics[width=\linewidth]{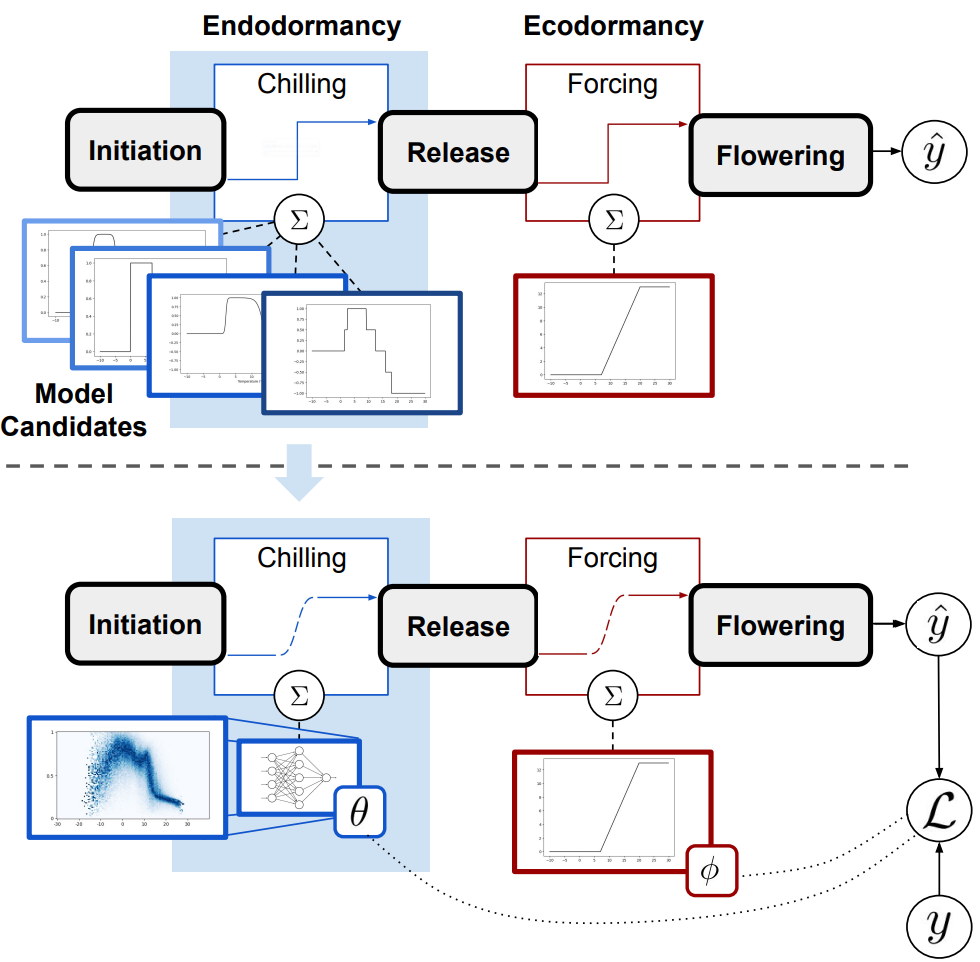}
    \caption{Schematic overview of two approaches modeling the phenological stages between dormancy initiation and flowering. Biophysical model candidates (Top) show structural discrepancies in how temperature contributes towards endodormancy progression. In this work (Bottom), we replace this module with an MLP and jointly optimize it with the biophysical model parameters, obtaining a new temperature response function. We then evaluate its ability to generalize to unobserved seasons and reflect on its biophysical plausibility. }

    \label{fig:main-figure}
\end{figure}

In this work, we propose a hybrid model for predicting phenological transitions by providing a generalized piece-wise differentiable approximation to commonly used biophysical models of tree dormancy and substitute the component showing large structural disagreement (i.e. the function relating daily temperatures to progression towards endodormancy release) with a multilayer perceptron (MLP), sketched in Figure~\ref{fig:main-figure}. Gradients with respect to the MLP parameters $\theta$ and process-based model parameters $\phi$ can be computed by performing backpropagation through the constrained model. These parameters are then jointly optimized using an extensive dataset on cherry tree flowering dates in Japan, Switzerland and South Korea, paired to MERRA-2 hourly temperature data. The substitution of the biophysical model component allows us to obtain a learned temperature response function provided by the MLP. We refrain from using more complex architectures, to enable direct inter-comparison of inputs/outputs with biophysical models that generally use stateless chill functions with a limited temporal context window (ruling out recurrent networks, for example). We compare the model to commonly used process-based models w.r.t. their ability to predict blooming dates. Finally, we reflect on the plausibility of the temperature response function learned by the MLP. 

\section{Methodology}\label{section:methodology}
We formalize the task of predicting blooming day-of-year (\doy) as a regression problem and compare models $M: \mathbb{R}^{S\times 24}\rightarrow\mathbb{R} $ that relate hourly temperature values for every day within the season (of length $S$) to a \doy\ value that is discretized during evaluation. 

\subsection{Dataset}\label{section:methodology:dataset} 
\begin{figure}[ht]
    \centering
    \includegraphics[width=\linewidth]{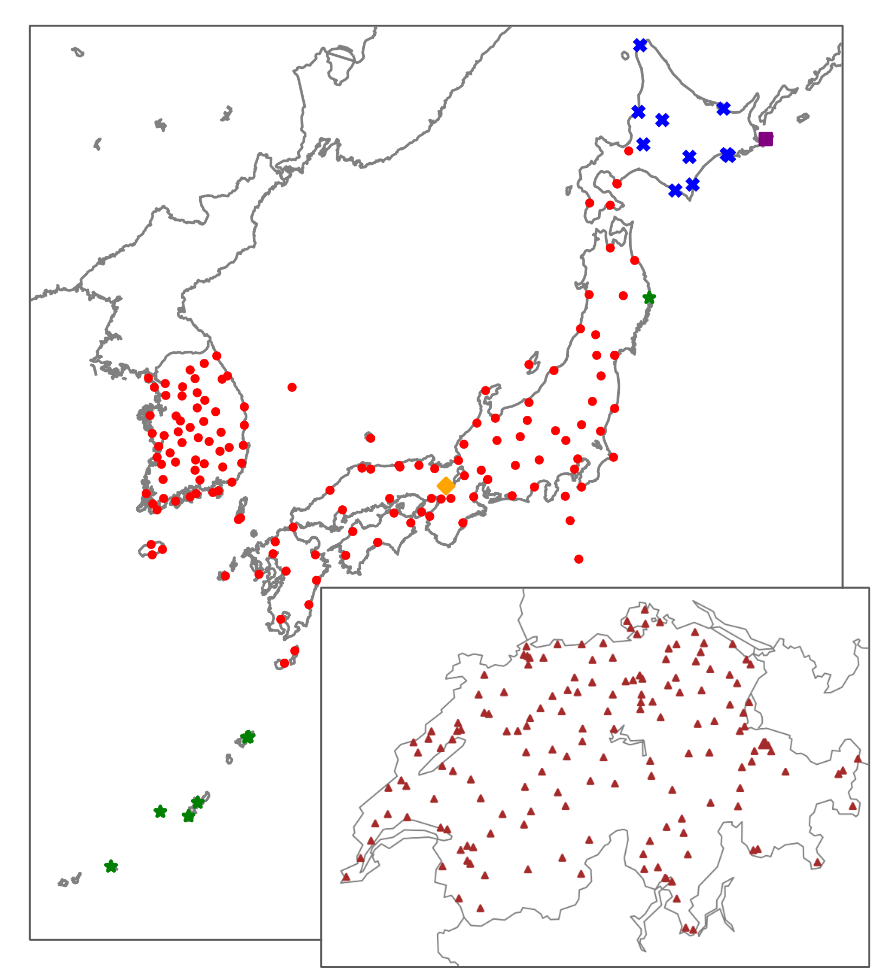}
    \caption{Spatial distribution of locations that were available in the dataset for each country (not to scale), colored by (partially assumed) tree variety occurrence. Varieties include: \textit{Prunus}\ \texttimes\textit{Yedoensis} (Red circles), \textit{Prunus} \textit{Sargentii} (Blue crosses), \textit{Prunus} \textit{Campanulata Maxim} (Green stars), \textit{Prunus} \textit{Nipponica Matsum} (Purple squares), \textit{Prunus} \textit{Jamasakura} (Orange diamonds), \textit{Prunus} \textit{Avium} (Brown triangles). }
    \label{fig:locations_varieties}
\end{figure}

The dataset used was based on data released by George Mason’s Department of Statistics as part of a competition on cherry blossom \doy\ prediction 
  \citep{gmu-competition}. It is composed of blooming dates for numerous locations with diverse climates scattered throughout Japan, Switzerland and South Korea. We paired observed dates $y$ in this dataset with MERRA-2 hourly surface temperature data $\mathbf{X}\in\mathbb{R}^{S\times 24}$ forming a dataset $\{(\mathbf{X}_n, y_n)\}_N$, where $\mathbf{X}_n=[\mathbf{x}_1, .., \mathbf{x}_S]$ contains the temperature data of the season matching flowering \doy\ $y_n$ in some year at some location. Similar to related work on modeling cherry tree phenology we set the 1st of October as start of the season and based on observed bloom \doy\ the season length was set to $S=274$ (ending on the 1st of July on non-leap years). MERRA-2 provides data starting from 1980, so any preceding blooming dates have been omitted. The locations were mapped to their respective tree variety as reported by the Japanese Meteorological Agency and Swiss Phenology Network for Japan and Switzerland, respectively. Not all tree varieties could be identified for the locations in Japan. These (7) locations were excluded. Trees in South Korea were assumed to be of the \textit{Prunus}\texttimes\textit{Yedoensis} variety based on previous work~\citep{hur2014change, chung2009using} and its frequent occurrence. 
Figure \ref{fig:locations_varieties} shows the locations and distribution of tree varieties included in the dataset. The process based models that are considered are not suitable for the subtropical regions (Okinawa, Amami) where the \textit{Prunus} \textit{Campanulata Maxim} trees grow. These were excluded to make a fair model comparison. Trees species that were represented in one only one location were removed as well. The Japanese dataset thus only contains data on the \textit{Prunus} \texttimes\textit{Yedoensis} and \textit{Prunus} \textit{Sargentii} varieties. Since the data sources for each country adhered to their own flowering definition, models are trained separately for Japan (N=3317), Switzerland (N=4780) and South Korea (N=930).

\subsection{Models}\label{section:methodology:models}

Three classes of models are evaluated in this work: mechanistic biophysical models, our proposed hybrid model, and purely data-driven neural networks. Since biophysical models show strong structural similarity, a generalized description is provided first, followed by an overview of how the different models fit within this framework. Subsequently, we describe our approach and finally the data-driven baseline.

\subsubsection{Phenology Models}\label{section:methodology:models:phenologymodelspreliminary}

\noindent Mechanistic phenology models of dormancy in deciduous trees generally consider two successive stages: endodormancy and ecodormancy. Before being able to transition between stages, a tree is required to spend a period in cold temperatures, which is typically modeled using the daily accumulation of temperature-dependent units (referred to as chilling units) that capture the progression towards completing this requirement. Subsequently, before flowering, a tree is required to spend a period under warm temperatures that are modeled in a similar fashion using the accumulation of ``forcing" units towards some required threshold, after which blooming occurs. \\\
\indent Let $c(\mathbf{x}_t)$, $f(\mathbf{x}_t)$ ($c,f: \mathbb{R}^{24}\rightarrow\mathbb{R}_{\geq0}$) denote functions that define the respective daily contribution of chilling and forcing based on hourly temperature data. The exact formulation of how these functions relate temperature levels to contribution to the phenological stages is, for now, left unspecified, since this is where the considered biophysical models show discrepancies. Let $C_t$ denote the cumulative sum of chilling units at day $t$ and $r^{(c)}_t \in \{0,1\}$ an indicator variable specifying whether a threshold $\beta^{(c)}$ has been met:
\begin{equation}\label{eq:units_cs_chill}
    C_t=\sum\limits_{\tau=1}^t c(\mathbf{x}_\tau) \quad\quad r_{t}^{(c)}= \mathds{1}(C_t\geq\beta^{(c)})
\end{equation}
The accumulation of forcing units is only considered after the chilling requirement is fulfilled. Its cumulative sum $F_t$ and forcing requirement $r_t^{(f)}$ are defined similarly:
\begin{equation}\label{eq:units_cs_force}
    F_{t}= \sum\limits_{\tau=1}^t f(\mathbf{x}_{\tau}) r_{\tau}^{(c)} \quad\quad r_{t}^{(f)}=\mathds{1}(F_t\geq\beta^{(f)})
\end{equation}
where $r_t^{(f)}$ indicates that flowering has occurred and $\beta^{(c)}, \beta^{(f)}$ are parameters of the model. 

\subsubsection{Baseline Phenology Models}\label{section:methodology:models:phenologymodelsbaseline}

The biophysical models that are considered all fit in the framework described in the previous section, but differ in their choice of $c$ and $f$. A rough description of commonly used chilling models is provided below, followed by those used for forcing. For details on the exact weighting of temperature levels we refer to their original publication. The Chill Hours model~\citep{bennett1949temperature, weinberger1950chilling} simply counts the number of hours spent between 0°C and some threshold temperature (usually 7.2°C). The Utah model~\citep{richardson1974model} weights the effectiveness of hourly chill unit accumulation based on the respective temperature values. The Chill Days model~\citep{cesaraccio2004chilling} shares a parameter $T_B$ that controls the optimal temperature for both chilling and forcing accumulation to be suitable for a wider variety of climates. Among these chilling models, only the Chill Days model was introduced with a complementary forcing function $f$. For the other models one was assigned instead. There is less variation between forcing functions and most models adapt a Growing Degree Days approach to hourly timesteps~\citep{richardson1975pheno}, where some apply an additional weight function~\citep{anderson1985validation, luedeling2021phenoflex}. In this work we follow this Growing Degree Hours (GDH) approach, without the additional weight function:
\begin{equation}\label{eq:def_gdd}
    f_{\text{GDH}}(\mathbf{x}_t)=\sum\limits^{24}_{i=1}\max(0, x_{t, i} - T_B)
\end{equation}
where $T_B$ is a base temperature parameter for controlling the effectiveness of temperature levels to forcing~\citep{richardson1974model}. In total, all models have three parameters which we collectively define as $\phi=(\beta^{(c)}, \beta^{(f)}, T_B)$.




\subsubsection{Proposed Model}\label{section:methodology:models:proposedmodel}
In this section we describe how the mechanistic phenology model framework is adapted to enable substitution of the chill response function. A (piece-wise) differentiable approximation to the mechanistic models is built using generalized logistic functions to approximate the step functions used by these models:
\begin{equation}\label{eq:soft_threshold}
    \sigma(x;\alpha,\beta, \gamma)=\frac{\gamma}{1+e^{-\alpha(x-\beta)}}
\end{equation}
where $\alpha$ controls the slope, $\beta$ the inflection point and $\gamma$ the right-hand side asymptote. Mechanistic models generally agree on the forcing function $f$ that is used. Moreover, it meets the piece-wise differentiability requirements (Eq. \ref{eq:def_gdd}) and thus remains unchanged. The chilling function $c$, however, shows stronger dissimilarities and is substituted by a learnable multilayer perceptron (MLP) consisting of two hidden layers with each containing 64 neurons and a ReLU activation function. In the final layer, a sigmoid function is used instead, such that $C_t$ is bounded between $0$ and season length $S$.

Let $\tilde{c}(\cdot; \theta)$ denote the daily MLP chilling contribution parameterized by $\theta$. The complete description of our model is then provided by 
\begin{equation}\label{eq:units_cs_chill_approx}
    \tilde{C}_t=\sum\limits_{\tau=1}^t \tilde{c}(\mathbf{x}_\tau; \theta) \quad\quad \tilde{r}_{t}^{(c)}= \sigma(\frac{\tilde{C}_t}{
S}; \alpha^{(c)}, \beta^{(c)}, 1)
\end{equation}
\begin{equation}\label{eq:units_cs_force_approx}
    \tilde{F}_{t}= \sum\limits_{\tau=1}^t f(\mathbf{x}_{\tau}) \tilde{r}_{\tau}^{(c)} \quad\quad \tilde{r}_{t}^{(f)}=\sigma(\frac{\tilde{F}_t}{S};\alpha^{(f)}, \beta^{(f)}, 1)
\end{equation}
Two additional parameters $\alpha^{(c)}, \alpha^{(f)}$ are introduced that control the slope of the threshold approximation. $\tilde{C}_{t}, \tilde{F}_{t}$ are divided by $S$ to limit the size of the exponent in Eq. \ref{eq:soft_threshold} for numerical stability. The original model structure, and thus parameters $\phi=(\beta^{(c)}, \beta^{(f)}, T_B)$, are preserved. \par

The soft indicator $\tilde{r}_{t}^{(f)}$ no longer provides a binary indication of whether flowering has occurred, and we assign it a probabilistic interpretation. That is, we consider $\tilde{r}^{(f)}_t=\sigma(\cdot;\alpha^{(f)},\beta^{(f)}, 1)$ to be the cumulative distribution function of a logistic distribution over $\frac{F_{t}}{S}$ defining the probability that blooming has already occurred at a certain stage of development (as provided by $\frac{F_{t}}{S}$). Similar to~\citep{terres2013analyzing} and~\citep{allen2014modeling}, we then define the probability of flowering at day $t$ and minimize the corresponding negative log-likelihood: 
\begin{equation}
    \mathcal{L}(t,\theta,\phi)=-\log(\tilde{r}^{(f)}_t-\tilde{r}^{(f)}_{t-1}).
\end{equation} 
During evaluation, \doy\ predictions are obtained by considering the day with maximum probability of blooming $\argmax\limits_t \lbrack \tilde{r}^{(f)}_t-\tilde{r}^{(f)}_{t-1} \rbrack$. 

\subsubsection{Data-driven Baseline}\label{section:methodology:models:lstm}

A data-driven baseline consists of a 1D convolutional neural network~\citep{fukushima1980neocognitron, lecun1989handwritten} (CNN), as well as a two-layer long-short term memory~\cite{hochreiter1997lstm} (LSTM) model with a hidden state size of 128. At every point in time, this hidden state is the input of a linear layer with a final sigmoid activation to obtain daily probability values. This results in a model size that is an order of magnitude larger than the hybrid model, but we found that this was required to generalize well in all settings. We follow the approach used by~\cite{saxena2023multi} and use a binary cross-entropy loss on the daily predictions provided by the model. When compared to models with location-specific parameters $\phi$, normalized coordinates are appended to the model input to be able to distinguish between locations as well. These coordinate features are omitted otherwise. 

\subsection{Model Optimization}\label{section:methodology:optimization}
\noindent\textbf{Mechanistic Models}\quad For each of the considered models, a parameter grid search was performed over three parameters: chill requirement $\beta^{(c)}$, forcing requirement $\beta^{(h)}$ and base temperature $T_{B}$. Parameters were optimized per location or tree variety and selected based on the mean squared error obtained using the training dataset.



  

\noindent\textbf{Hybrid Model}\quad The differentiability of the model allows us to perform backpropagation to compute gradients w.r.t. both the biophysical model parameters $\phi$ and the MLP parameters $\theta$. Gradients were computed using PyTorch~\citep{paszke2019pytorch} Autograd over the complete training dataset. An Adam optimizer~\citep{kingma2014adam} was used to update the model parameters with an initial learning rate of $1e^{-3}$ and weight decay of $1e^{-4}$. Over a total of 20000 epochs this learning rate decayed by a factor of 0.9 for every 2000 iterations over the dataset. Chilling requirement slope $\alpha^{(c)}$ was set to 50 to closely approximate the step function used in process based models. Forcing requirement slope $\alpha^{(f)}$ was interpreted as the scale parameter of a logistic distribution. Its value was initialized to 1 and learned during the training procedure, while enforcing a lower bound of 0.01. \par
\noindent\textbf{Data-driven}\quad The settings are identical to that of the hybrid model, except that, due to memory limitations, gradients are computed in batches with a batch size set to 512. \par

%
%
\section{Experiments}\label{section:results}

To investigate the generalization capabilities of each model, they are fit and evaluated in several settings, as described below. Every experiment was run 10 times using different random seeds. Table~\ref{tab:model_evaluation_scores} reports the mean absolute error (MAE) of predicted \doy\ blooming dates, as well as its standard error, obtained for each setting. MAE was chosen over other metrics for its lower sensitivity to outliers, since we found that the mechanistic models were more prone to give end-of-season predictions if thresholds were not met.

\begin{table}[ht]
    \centering
    \begin{adjustbox}{width=\linewidth,keepaspectratio}
    \begin{tabular}{|l|c|c|c|}\hhline{|=|===|}
         & \multicolumn{3}{|c|}{\textbf{MAE \textpm SE}} \\\hhline{|=|===|}
        
        & \multicolumn{3}{|c|}{\textbf{Setting 1 - Temporal}} \\
        \textbf{Model} & \textbf{Japan} & \textbf{Switzerland} & \textbf{South Korea} \\\hline
        Chill Hour & \ 3.76 \textpm 0.11 & \ 6.52 \textpm 0.18 & \ 6.42 \textpm 0.36 \\
        Utah Chill & \ 3.07 \textpm 0.06 & \ 7.32 \textpm 0.28 & \ 6.12 \textpm 0.27  \\
        Chill Days & \ 3.71 \textpm 0.09 & \ 8.47 \textpm 0.29 & \ 7.13 \textpm 0.44  \\
        \modelname & \ \textbf{2.17 \textpm 0.03} & \ \textbf{5.36 \textpm 0.15} & \ \textbf{4.86 \textpm 0.06} \\
        CNN  & \ 3.80 \textpm 0.09 & 11.28 \textpm 0.17 & \ 6.21 \textpm 0.17 \\   
        LSTM & \ 2.80 \textpm 0.06 & \ 7.11 \textpm 0.09 & \ 7.22 \textpm 0.57 \\\hline 
        
                & \multicolumn{3}{|c|}{\textbf{Setting 2 - Temporal (Variety)}} \\
        \textbf{Model}  & \textbf{Japan} & \textbf{Switzerland} & \textbf{South Korea} \\\hline
       Chill Hour & 18.03 \textpm 0.28 & 17.64 \textpm 0.35 & \ 9.50 \textpm 0.26 \\
       Utah Chill & 15.65 \textpm 0.31 & 18.14 \textpm 0.34 & \ 6.97 \textpm 0.17  \\ 
       Chill Days &  16.68 \textpm 0.30  & 18.54 \textpm 0.43 & 10.76 \textpm 0.54 \\
       \modelname & \ 3.28 \textpm 0.04 & \textbf{10.2 \textpm 0.15} & \ \textbf{5.41 \textpm 0.08}  \\
       CNN & \ 3.98 \textpm 0.10 & 12.21 \textpm 0.31 & \ 6.21 \textpm 0.13 \\
       LSTM & \ \textbf{3.24 \textpm 0.05} & 12.75 \textpm 0.35 & \ 7.10 \textpm 0.37  \\\hline
       
                & \multicolumn{3}{|c|}{\textbf{Setting 3 - Spatiotemporal (Variety)}} \\
        \textbf{Model}  & \textbf{Japan} & \textbf{Switzerland} & \textbf{South Korea} \\\hline
       Chill Hour & 18.34 \textpm 0.78 & 17.44 \textpm 0.62 & 11.23 \textpm 0.98 \\
       Utah Chill & 15.29 \textpm 0.73 & 18.13 \textpm 0.59 & \ 7.31 \textpm 0.55 \\
       Chill Days & 16.80 \textpm 0.96 & 18.10 \textpm 0.69 & 13.62 \textpm 2.15 \\
       \modelname & \ \textbf{3.64 \textpm 0.09} & \textbf{10.6 \textpm 0.37} & \ \textbf{6.07 \textpm 0.38} \\
       CNN  & \ 4.09 \textpm 0.12 & 12.66 \textpm 0.34 & \ 7.00 \textpm 0.43 \\
       LSTM & \ 3.94 \textpm 0.18 & 13.24 \textpm 0.41 & \ 8.30 \textpm 0.94 \\\hline

    \end{tabular}
    \end{adjustbox}
    \caption{Mean Absolute Error (in days) and Standard Error obtained from evaluating three process-based models (Chill Hour, Utah Chill and Chill Days models), an LSTM, CNN, as well as the proposed model (\modelname) in three different countries on a held out test dataset. The models are evaluated for their generalization capabilities in multiple settings, as described in the results section.}
            

    \label{tab:model_evaluation_scores}
\end{table}

\subsubsection*{1. Temporal Generalization per Location}
The first setting focuses on the ability of the models to predict blooming dates in years that are not used to fit the model, and for the same locations used for training. The years that span the dataset (1980-2021) are randomly split in train ($75\%$) and test years to evaluate the model in unobserved weather conditions. No restrictions are imposed on the parameters $\phi$, allowing them to take on different values for each considered location. 
Table~\ref{tab:model_evaluation_scores} shows the ability of the three biophysical models to generalize to new years based on the parameters obtained using grid search. These models show comparable results in predicting blooming dates within each of the countries. All models are best able to predict blooming dates in Japan, followed by South Korea and Switzerland. We suspect this is caused by the lower data availability in South Korea with fewer data points available per location and stronger influence of local temperature differences in Switzerland due to elevation/shading. 
The LSTM, on the other hand, performs comparably in South Korea and Switzerland in the temporal generalization setting, since the noisier nature of the Swiss dataset may be compensated by a higher number of training samples compared to South Korea. In this lower data regime, the process-based models tend to outperform the LSTM. On the other hand, the LSTM outperforms all process-based models in data-rich Japan.

From Table \ref{tab:model_evaluation_scores} it can be seen the proposed model shows a lower mean absolute error than all process-based models, as well as the LSTM, in each of the considered countries when evaluated on data from held out test years. Interestingly, our hybrid model behaves akin to the process-based models in that it is able to fit better to the South Korea dataset than in Switzerland. At the same time, it is able to profit from the higher data availability in Japan, thus outperforming all baselines on all datasets. 

\noindent The proposed model is constructed using two adaptations from mechanistic models, namely the piece-wise differentiable approximation of the model structure and the substitution of the chill temperature response function by an MLP. To investigate whether it is truly the learned chill function that causes the improvement in model performance, when compared to traditional models, and not the differentiable model approximation, we optimize this approximation when using the Utah chilling function instead. The resulting MAE of $3.17$ \textpm $0.07$ days for Japan does not match the 2.17 \textpm 0.03 that was obtained when using our hybrid model, indicating the learned chill temperature response function does indeed cause the model improvement. 

\subsubsection*{2. Temporal Generalization per Variety}

\begin{figure*}[ht]
    \centering
    \renewcommand{\arraystretch}{.7} 
    \begin{tabular}{|c|c@{\hskip3pt}c|c|c|}\hline
         & & & & \\
         & & Temporal Data Split & Temporal Data Split & Spatiotemporal Data Split \\
         & & & (per Variety) & (per Variety) \\\hline
         & & & & \\
        \parbox[t]{2mm}{\rotatebox[origin=l]{90}{\hspace{1em}Utah Chill}} & \parbox[t]{2mm}{\rotatebox[origin=l]{90}{\hspace{1em} Observed DOY}} & \includegraphics[width=.25\linewidth, trim={10mm 10mm 0 0},clip]{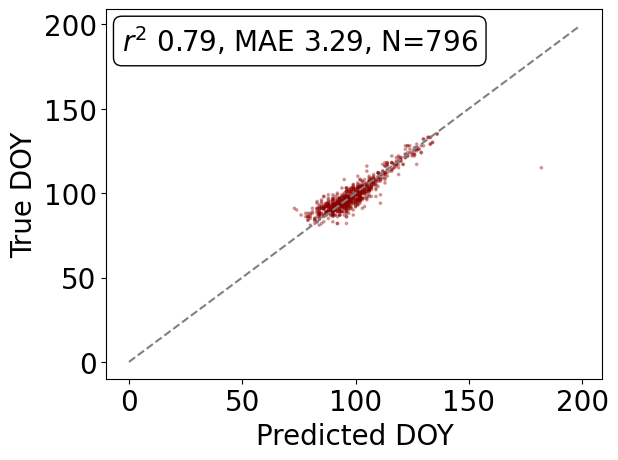} & \includegraphics[width=.25\linewidth, trim={10mm 10mm 0 0},clip]{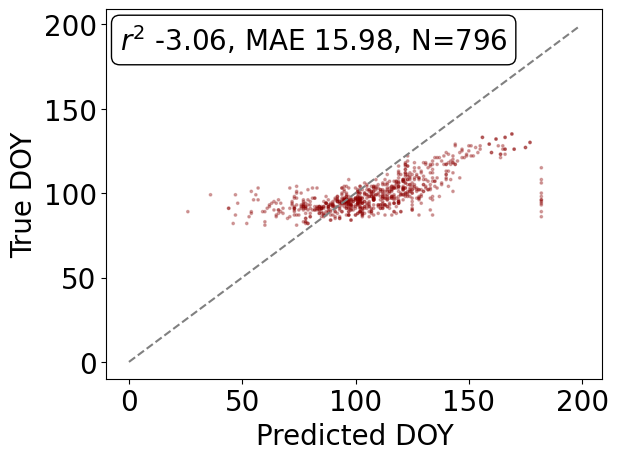} & \includegraphics[width=.25\linewidth, trim={10mm 10mm 0 0},clip]{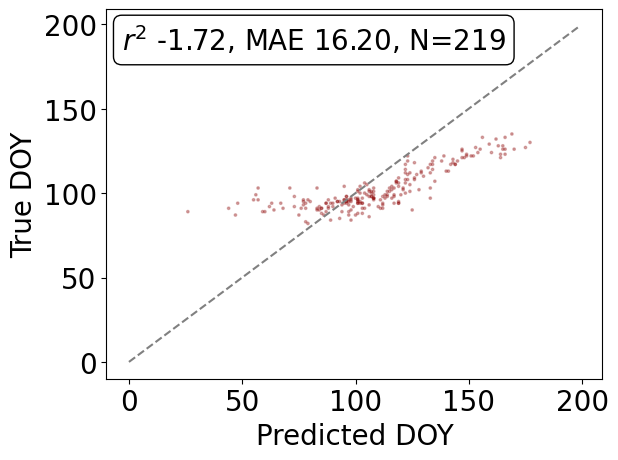} \\
        \parbox[t]{2mm}{\rotatebox[origin=l]{90}{\hspace{2em}Hybrid}} & \parbox[t]{2mm}{\rotatebox[origin=l]{90}{\hspace{1em} Observed DOY}} & \includegraphics[width=.25\linewidth, trim={10mm 10mm 0 0},clip]{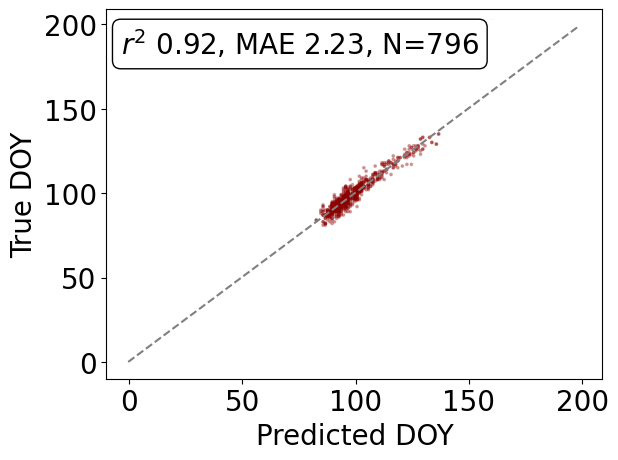} & \includegraphics[width=.25\linewidth, trim={10mm 10mm 0 0},clip]{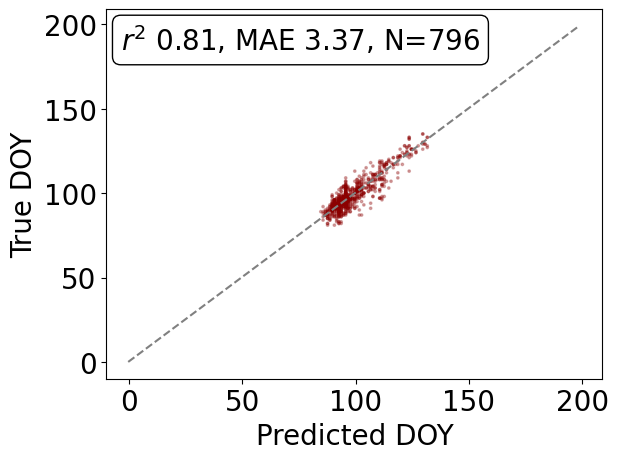} & \includegraphics[width=.25\linewidth, trim={10mm 10mm 0 0},clip]{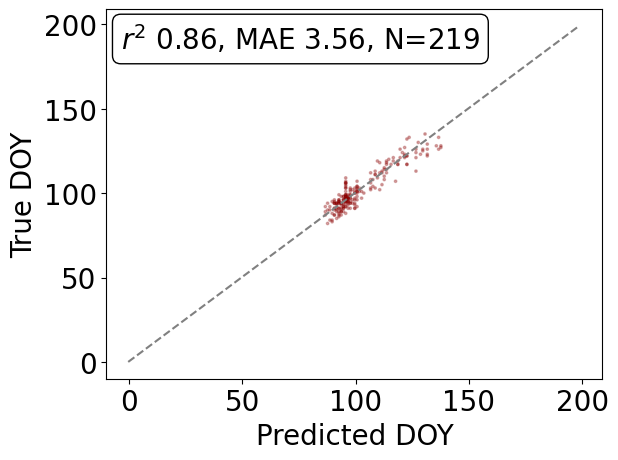} \\
        & & Predicted DOY &  Predicted DOY & Predicted DOY \\\hline
    \end{tabular}
    \caption{Scatter plots of blooming date predictions for all locations observing the \textit{Prunus} \texttimes\textit{Yedoensis} tree variety in Japan. Plots are shown for the Utah model (Top) and the model with learned chilling function (Bottom) in the three evaluation settings. It can be seen that, once consistent tree variety parameters are enforced in all locations the mechanistic model underfits, whereas the temperature response function learned by the hybrid model transfers to many locations. }
    \label{tab:scatter_plots_pb_df_japan_settings}
\end{figure*}

A common use-case for phenology models is to be applied in locations without data availability, and model parameters are thus often identified per tree variety, rather than per location. To investigate our model's ability to learn such parameters, we use an evaluation setup identical to the first with the exception that models are calibrated per tree variety (Figure~\ref{fig:locations_varieties}), rather than having separate parameters $\phi$ for each location. This results in all locations with the same tree variety sharing parameters and thus showing the same response to temperature. 
In this case, Table~\ref{tab:model_evaluation_scores} and Figure~\ref{tab:scatter_plots_pb_df_japan_settings} show that the biophysical models underfit, and their ability to predict blooming dates degrades. The assumption is especially restrictive in countries with numerous locations over different climatic zones, as can be seen by the stronger model degradation in Japan and Switzerland. The expressiveness of the LSTM prevents it from underfitting in this setting and, with sufficient data, such as in Japan, it shows the lowest MAE of all models. In Switzerland and South Korea, this benefit is not apparent. 
Our hybrid model matches the LSTM in its generalization capability in Japan and outperforms all three biophysical models and LSTM in Switzerland and South Korea. The imposed structure helps in settings with less available data or data of lower quality, while its data-driven nature allows generalization across locations. 





\subsubsection*{3. Spatio-temporal Generalization}
A subsequent step is to evaluate whether the variety-specific model parameters actually generalize to unseen locations as well. To do this, we hold out $25\%$ of locations to be used as test, while keeping a temporal train/test split as in the previous settings. Variety-specific parameter sets are learned using the training dataset in the same manner as in the previous setting. 
Based on the results of the previous evaluation setting, our hypothesis was that the biophysical model parameters do not generalize well to held out locations. Surprisingly, the overparameterized LSTM manages to retain much of its generalization capability in Japan. The hybrid model, however, manages to improve over all models with a lower increase of MAE (roughly half a day) for each country. 

We expect the temporal train/test split to mitigate the effects of spatial autocorrelation between samples in the two datasets that may bias the evaluation. However, the climates represented in both sets will still be relatively similar, and we expect the location generalization capabilities to decrease the further its climate deviates from the training distribution.


\begin{figure*}[ht]
    \centering
    \begin{tabular}{|c|ccc|}\hline
        \parbox[t]{2mm}{\multirow{4}{*}{\rotatebox[origin=c]{90}{\textbf{Hybrid}\hspace{4em}}}} & Seed 1 & Seed 2 & Seed 3 \\
        & \includegraphics[width=.28\linewidth]{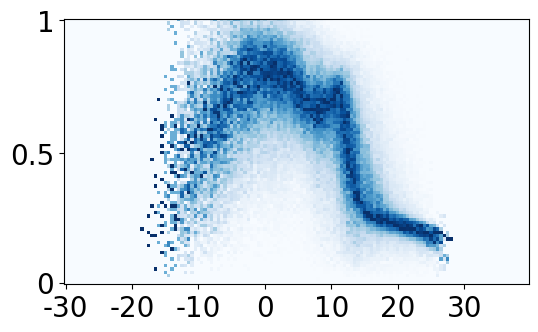} & \includegraphics[width=.28\linewidth]{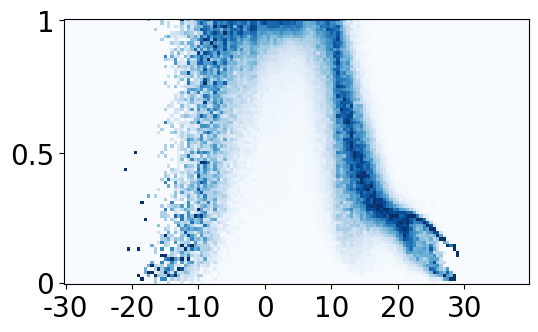} & \includegraphics[width=.28\linewidth]{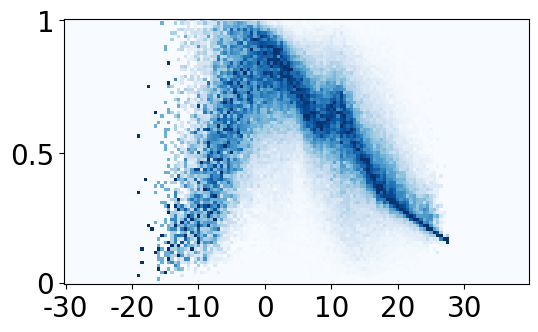} \\\hline
        
        
        \parbox[t]{2mm}{\multirow{2}{*}{\rotatebox[origin=c]{90}{\textbf{Biophysical}\hspace{2.5em}}}} & Chill Hours Model & Utah Chill Model & Chill Days Model \\
        & \includegraphics[width=.28\linewidth]{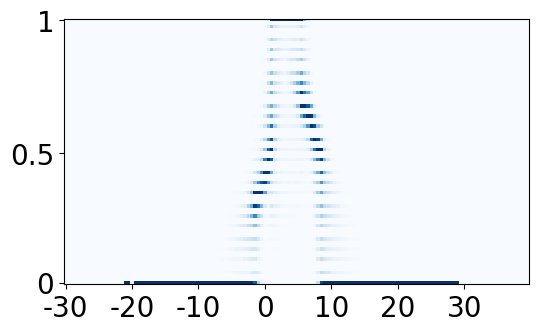} & \includegraphics[width=.28\linewidth]{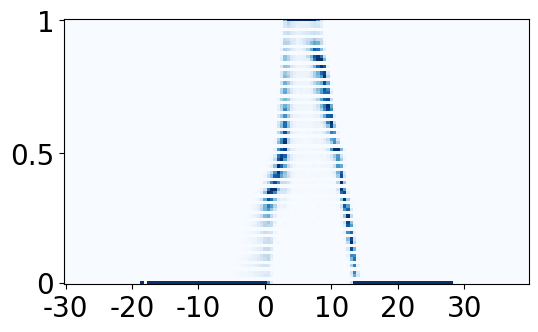} & \includegraphics[width=.28\linewidth]{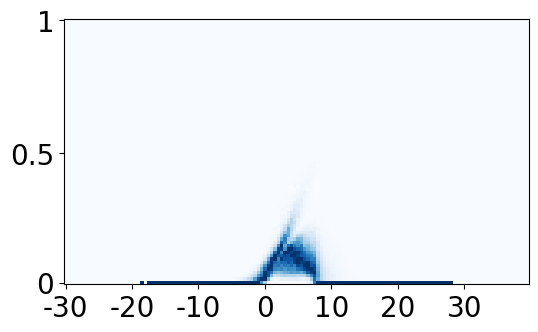} \\\hline
        \end{tabular}
    \caption{(Top) Visualization of the chill response function learned by the MLP in Japan for the first three random seeds used for evaluation, where the seed controls MLP weight initialization as well as the train/test split. Plots showing the effect of each source of randomness are included in the attached repository. The figure shows the intensity of the response plotted against the mean of the function input (i.e. mean temperature in \textdegree C of the 24 hourly measurements of the respective day) for all days occurring in the test dataset.  Darker colors highlight the more frequently observed output at each temperature level by showing the density of each column (temperature level discretized to intervals of 0.5 \textdegree C). (Bottom) Same visualization when made using the three biophysical models used as baselines. The biophysical models show striped artifacts due to their weighting of discrete temperature intervals.}
    \label{tab:temp_resp_func}
\end{figure*}

\subsubsection*{Visualizing the Chill Response Function}
An approximation of the learned chill function is depicted in Figure \ref{tab:temp_resp_func} by plotting the mean daily temperature and corresponding chill response function output density for all samples in the test set in Japan. The color intensity reflects the density of a response value within each column in temperature intervals of 0.5 \textdegree C. Within these temperature levels, some variation in the chill response can be seen, although usually clustered in a small interval. All learned functions show a large variation in output in the colder temperature levels (roughly between -20 to -5 \textdegree C), perhaps due to these being less represented in the training dataset. All functions show a strong response around the same temperature range (-5 to 15 \textdegree C) with peaks mostly occurring around 0\textdegree C. Chill contribution decreases strongly in warmer temperature levels, and all levels provide some non-zero contribution towards fulfilling the chill requirement. In Switzerland and South Korea, the model does not learn smooth functions weighting temperature contribution but shows a binary response only giving an output of 1 at temperatures below 15 \textdegree C, indicating the importance of the dataset quality. 

\noindent Experiments were executed on a Lenovo p16 laptop (Intel Core i9-12950HX, 2300Mhz, 32GB RAM, RTX A5500 GPU) running Windows 10 Enterprise. All code used to obtain the publicly available datasets and run the experiments is available on GitHub.


%
%
\section{Discussion}\label{section:discussion}

The proposed hybrid model shows improved generalization capabilities compared to both the mechanistic models and  LSTM. When data is limited, such as in South Korea, LSTM models tend to overfit to the available data and have difficulty to match the performance of much simpler biophysical models. The approximations provided by these models, however, are not expressive enough and thus underfit in settings with high data availability, as is the case in Japan. The additional flexibility of the proposed hybrid approach with respect to the biophysical models learns an approximation that, within its structural constraints, obtains a closer fit and improves generalization within all the evaluated settings. Moreover, for the widely occurring \textit{Prunus} \texttimes\textit{Yedoensis} variety, a parameter set was obtained that captures the tree behaviour well in a broad range of environments (Figure \ref{fig:locations_varieties}), enabling its usage in locations with low data availability if the variety is known. The mechanistic models, however, are not expressive enough to do the same and require re-calibration per location. 


However, there is no guarantee that the hybrid model, within its biophysical constraints, learns a function that actually resembles the biological processes behind the phenological changes, since it is solely optimized for prediction under the provided training dataset. In fact, Figure~\ref{tab:temp_resp_func} shows that, for different random weight initializations and dataset splits, different functions are obtained.
Moreover, the hybrid model displays a wider range of temperatures that contribute to phenological development than all biophysical models. The Utah model~\citep{richardson1974model}, for example, originated from the following assumption: \textit{``The chilling effect that advances rest completion has an optimum at 6\textdegree C and is lost at 0\textdegree C and 12.5\textdegree C."} The learned chill function does not show consistent peaks around 6\textdegree C but does show increased contribution between the 0\textdegree C - 10\textdegree C temperature range. Chill effect is not lost outside 0\textdegree C - 12.5\textdegree C. Subzero temperatures do show less contribution as temperatures become colder but with large variation in effect. A decreasing trend is also observed on the other extreme, where chilling effects become less significant in warmer days, but never reaching a level of zero effect. Here, the model conditionally shows implausible behavior, since none of the biophysical models show chill contribution above 12.5\textdegree C.

Despite obtaining different response functions for different weight initializations, Table \ref{tab:model_evaluation_scores} shows little variation in their ability to predict blooming dates, indicating the function is underconstrained. Subsequent research could focus on whether enforcing biophysical resemblance through regularization or including more observations can be used to obtain a temperature response function with more similarity to existing models without sacrificing generalization ability. In particular, extreme weather events are underrepresented in the training data and no observations include the possibility of not flowering. In this way, the proposed model utilizes the predictive ability of machine learning methods while allowing direct comparison with mechanistic models through its biophysical constraints, therewith providing a promising avenue for improved phenology modeling.

\section{Conclusion}\label{section:conclusion}
In this work, we demonstrate that machine learning models can help biophysical models (and vice versa) for improved predictions on plant phenology by substituting a component modeling temperature contribution towards dormancy by a learnable function. 
In all considered environments, this hybrid model improves with respect to three commonly used mechanistic models that were fit using an exhaustive grid search. The structural approximation provided by the mechanistic models underfits on the data, which becomes more apparent when learning a single set of parameters for all locations with the same tree species/variety. Under conditions with sufficient data quality and quantity, the flexibility of the learnable component allows the hybrid model to fit these parameters, while retaining generalization capabilities. Preliminary inspection of the learned chill function shows resemblance to biophysical models, although it provides no guarantee for biophysical plausibility. Indeed, some variation between the temperature response can be seen when training the model with different random seeds, although the overall behaviour stays consistent. Despite this, each variation fits equally well to the data, indicating the model remains under-constrained. Further work could investigate whether scaling the model to additional environments results in less variation between the learned models. Alternatively, regularization of the chill function could enforce biophysical plausibility without affecting its generalization ability.

%
%
\section{Acknowledgements}
We would like to thank H.A. Baja, D.R. Paudel, L. Sweet and M.C. Ru{\ss}wurm for their comments and the discussions when reviewing this work. We thank the organizers of the George Mason University International Cherry Blossom Prediction Competition \citep{gmu-competition} for compiling an extensive dataset on cherry tree phenology. This work was partially supported by the Horizon Europe project PHENET - Tools and methods for extended plant phenotyping and envirotyping services of European research infrastructures (Grant agreement ID 101094587). 

%
%

\bibliography{references.bib}

\end{document}